\documentclass[conference]{IEEEtran}
\IEEEoverridecommandlockouts
\usepackage{cite}
\usepackage{amsmath,amssymb,amsfonts}
\usepackage{algorithmic}

\usepackage[ruled,vlined]{algorithm2e}
\usepackage{siunitx}
\SetArgSty{}

\usepackage{multirow}
\usepackage{graphicx}
\usepackage{textcomp}
\usepackage{xcolor}
\usepackage{booktabs}
\usepackage{colortbl}
\usepackage{tabularx}
\usepackage{booktabs}
\usepackage{ragged2e}
\usepackage{microtype}
\usepackage{amsmath}

\def\BibTeX{{\rm B\kern-.05em{\sc i\kern-.025em b}\kern-.08em
    T\kern-.1667em\lower.7ex\hbox{E}\kern-.125emX}}
\begin{document}

\title{\textbf{\textit{Towards Robust and Transferable IIoT Sensor based Anomaly Classification using Artificial Intelligence}} \huge \\ 
}

\author{\IEEEauthorblockN{Jana Kemnitz}
\IEEEauthorblockA{Siemens Technology}
\and
\IEEEauthorblockN{Thomas Bierweiler}
\IEEEauthorblockA{Siemens Digital Industries}
\and
\IEEEauthorblockN{Herbert Grieb}
\IEEEauthorblockA{Siemens Digital Industries}
\and
\IEEEauthorblockN{Stefan von Dosky}
\IEEEauthorblockA{Siemens Digital Industries}
\and
\IEEEauthorblockN{Daniel Schall}
\IEEEauthorblockA{Siemens Technology}


}

\maketitle

\begin{abstract}
The increasing deployment of low-cost industrial IoT (IIoT) sensor platforms on industrial assets enables great opportunities for anomaly classification in industrial plants. The performance of such a classification model depends highly on the available training data. Models perform well when the training data comes from the same machine. However, as soon as the machine is changed, repaired, or put into operation in a different environment, the prediction often fails. For this reason, we investigate whether it is feasible to have a robust and transferable method for AI based anomaly classification using different models and pre-processing steps on centrifugal pumps which are dismantled and put back into operation in the same as well as in different environments. Further, we investigate the model performance on different pumps from the same type compared to those from the training data.
\end{abstract}


\section{Introduction}

The NAMUR open architecture (NOA) enables the monitoring and optimization sensors of existing "brownfield" plants in the process industry. It sketches a second data channel in addition to existing core process control systems like Simatic PCS neo. With the help of “low cost multi-sensors”, previously non instrumented assets can be retrofitted with a communication layer. This enables monitoring and classification of the operational states and anomalies of an asset based on the retrofitted sensor measurements. 
Machine learning models have great potential for these classification tasks for each individual asset in a production plant. The performance of machine learning models, however, strongly depends on the available training data and respective data distribution. Acquiring training data and labels through measurements of normal and anomaly conditions of industrial assets is expensive and very time-consuming. Anomaly conditions may negatively impact these monitored assets by down wearing or wrecking. Further, it is often difficult to transfer the models from one asset to another one, even if those are from the same type and highly standardized. Numerous real-world factors, as minimal divergence in production, wear and tear, or environmental factors influences the data distribution recorded by a sensor and consequently the derived predictions of machine learning models. Further, the data distribution is influenced by the sensor itself with several degrees of freedom in position and rotation. On the other hand, the same, industry-standard assets are often used in production facilities around the world. For example, having hundreds of the same type of filling pumps in one food and beverage plant is very common. Therefore, a robust and reusable model for a specific asset type is required for an application of economic condition monitoring. In other words, a single robust machine learning model, which can be delivered together with a respective sensor. 

Therefore, aim of this paper are the following: 
(i) introduce the IIoT measurement system and deployment, 
(ii) analyze challenges deriving requirements for a robustness and scalable  model dissemination, 
(iii) systematically evaluate the impact and challenges of production divergence, wear and tear, and maintenance in the context of robustness and transferability in machine learning models 
(iv) compare a feature-based Neural Network approach and ROCKET, including different pre-processing and post-processing combinations 
(v) derive initial guidelines to address these complex classification problem.

\section{Related Work}
\subsection{IIoT Sensor Systems} In the era of industry 4.0, Industrial IoT (IIoT) sensor devices are increasingly used to monitor and adapt to changes in the environment  \cite{Tong2019, Zhao2017, Schneider2018}. Sensors can capture a variety of physical values as light, temperature, pressure, vibration, and sound. The information is linked to the IIoT network to share data and connect between devices and management systems.  The correct operation of IIoT sensors play an essential role in the overall system performance \cite{Lee2015}. Several IIoT sensor kits exist \cite{Tong2019, Zhao2017}, but only a few are appropriate for industrial conditions. In the industrial context, IIoT sensor devices are often deployed in harsh environments, with ambient temperature, humidity, and strong vibrational conditions \cite{Gungor2009}. Further, a simple communication and a long battery and overall sensor lifespan is required \cite{Gungor2009}. Additionally, sensors within the IIoT should be high quality and low cost so that they can be used in very large numbers and enable the data collection from the variety of physical values simultaneously. The SITRANS multi sensor was specifically developed for industrial applications and its requirements \cite{Bierweiler2019}. Further, the compromise between high sampling rate and long sampling duration required for an accurate model prediction and limited data acquisition required for a long battery lifespan still remains an open challenge \cite{Schneider2018}. 

\subsection{Time Series Classification} Several traditional and deep learning based approaches for time series classification have been explored over the last decade \cite{IsmailFawaz2019, Dempster2020, Wang2017, Karim2018}. Often methods are tested on the UCR/UEA archive, an open collection of over 150 different data sets, having almost 20 sensor data sets available, with currently no data set in the automation or IIoT context. Deep learning approaches such as Convolutional Neural Networks (CNNs) and deep Residual Networks are less computational expensive and showed to be the most promising in a recent systematic method study  on the UCR/UEA data set \cite{IsmailFawaz2019, Wang2017, Karim2018}. In a recent paper \cite{Dempster2020}, ROCKET a simple linear classifier based on random convolutional kernel transformations showed a comparable high accuracy to Neural Networks (NN) at the UCR/UEA data set, but required only fraction of the computational expense compared to existing methods \cite{Dempster2020}. While, in general, these methods show great potential for time series classification, they have only been developed and tested within public repositories as the UCR/UEA archive \cite{Bagnall}. Besides the impressive number and amount of data collected, the transfer and robustness towards similar assets and the influence of maintenance work or environmental conditions have not been considered yet. However, it is important to evaluate whether a model dissemination from a lab to a real-world scenario can be achieved with the method developed.

\section{IIoT Measurement System}

\subsection{Industrial Asset}
An industrial asset can be any kind of machine or mechanical device that uses power to apply forces and control movement to perform an intended action. This asset can vary in size, purpose, and placement condition. Typical examples are pumps, motors, or manufacturing robots. An asset can be placed in a lot of scenarios: a motor pump combination can be placed outdoors or be part of a large bottling plant. 

\subsection{Sensor Measurement System}
With the help of low-cost multi-sensors measurement system, previously non instrumented assets can be retrofitted with a dedicated communication channel (Fig 1). These sensors can record airborne, structure-borne sound or temperature data, for example. We suggest that condition and the future behavior of any assets can be assessed using system-specific or central functions and methods (advanced analytic, scheduling). The respective experiments in this context have taken place only with the structure-bone data. These data offer a sampling rate of 6644 Hz. In order to keep the sensor battery lifespan as long as possible, the data is only collected for 512 samples every min, i.e. a total of 77 ms. Variations in the signal are observed due to any variations in the asset, the environment conditions of the asset, the asset health status and due to the sensor mounting and rotation. The sensor uses a Bluetooth low energy (BLE) wireless interface to communicate with a gateway device. Data collection is performed periodically for a short period of time. Within a fraction of a sec, the 512 samples are collected. Afterwards, the sensor goes into sleep mode, thereby saving energy. The gateway is connected to the internet and transmits the data via the Message Queuing Telemetry Transport (MQTT) protocol to the cloud-based backend services. Both, the number of samples and the data collection interval can be adjusted depending on the monitoring requirements. These requirements depend on the actual operational behavior of the asset.

\begin{figure}[ht]
	\centerline{\includegraphics[width=0.45\textwidth]{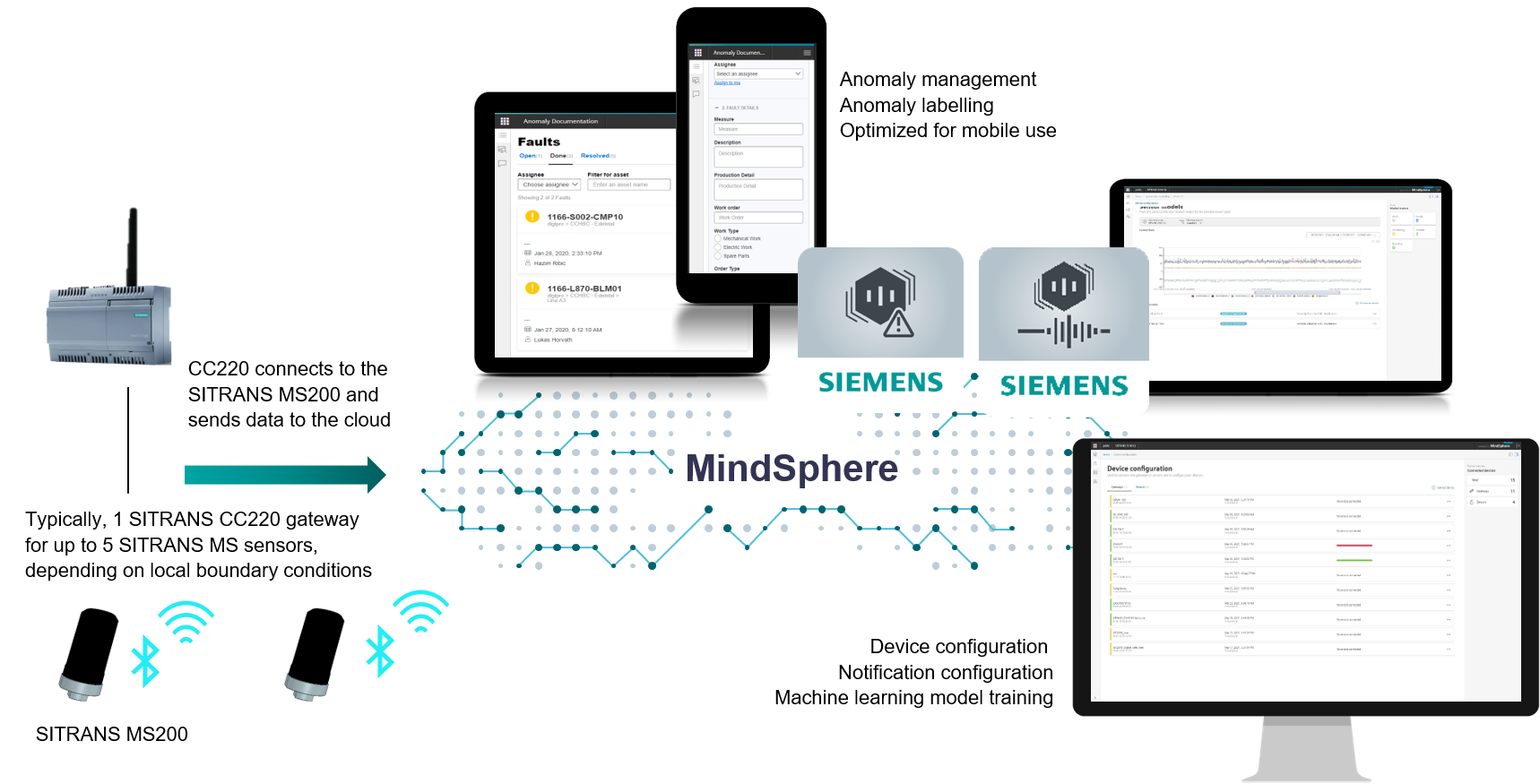}}
	\caption{IIoT measurement system and deployment}
	\label{fig1}
\end{figure}

\subsection{Machine Learning System and Deployment}
\subsubsection{Training}
The raw sensor data is stored in a blob-store in the cloud. The user can use a web dashboard to load the sensor data for a particular asset and sensor. The web dashboard is used by the machine operator to visually analyze and label the data. The dashboard offers a wizard-like approach to select the labelled data set and use a machine learning template to trigger the training. A template is a predefined machine learning algorithm and a set of hyperparameters that can be fine-tuned. The template abstracts the details of the algorithm so that non-ML experts can use the system. The actual training is done on a high-performance compute cluster. The cluster offers state-of-the-art elastic scaling capabilities, which is an essential requirement for large-scale sensor deployments. The final model is saved in the blob store. Model metrics are saved in a model lifecycle management database and can be visualized in the dashboard.

\subsubsection{Prediction}
The user can view the metrics and perform the model deployment. The system loads the model from the blob store and provisions the model to a runtime service called inference server. The inference server handles the execution of multiple concurrent models by subscribing to live sensor data in the cloud and passing the data to corresponding models. Pre- and post-processing steps are described at a later point in the paper.
Prediction results are handled by a rule-based system. For example, a detected anomaly is routed to the notification application. The user loads an anomaly documentation application to view and validate anomaly detection or classification results.

\subsection{Model Dissemination and Requirements}
Migrating a data science model from a research lab to a real-world deployment is non-trivial and potentially a continuous, ongoing process. Consequently, many machine learning models never go into production. A major challenge in industrial setups is the positive economic impact of a machine learning model. The economic impact can be considered positive when the costs and effort to create, deploy and update the model are below the savings gained by the model. Therefore, it is important to have a scalable machine learning solution that can be used on the same asset types in various environments. A scalable model can be considered as a digital product where replications are easy to achieve and can broadly distributed to the same asset-sensor combination anywhere in the world. This would result in high costs for the initial phase and marginal costs for each additional model application, mainly driven by the deployment infrastructure and low-cost sensors. 

A machine learning model can be considered as scalable when replications are easy to achieve and identical copies can broadly be distributed to the same asset-sensor combination anywhere in the world. This can be achieved in the two main machine learning paradigms, supervised and unsupervised learning, in different ways. In an unsupervised machine learning approach, the collection of training data is mostly automated, does not require any labels and is consequently time and cost wise inexpensive. However, unsupervised learning can only be used in specific anomaly detection tasks. In the supervised learning approach, the training data requires labels, which are very hard to collect, especially in classes reflecting anomaly behavior. Therefore, in a supervised learning approach, scalability is achieved by an increased robustness of the machine learning model, so it can be transferred easily to any other asset. Our aim is to thrive forward towards an “all-in-one solution” machine learning model solution applicable for one asset type sensor combination. Therefore, we suggest employing the advantages of both, supervised and unsupervised learning through an ensemble based model voting of an anomaly classification and an anomaly detection model. 

Further, we believe that a stable prediction may be more important compared to an instant prediction. This reasoning is derived by the assumption, that in real-world applications, a normal or anomaly class of an industrial asset will be given through a longer period. Therefore, a smoothing filter for the resulting model probabilities is suggested. Another important point is to lower the entry barrier as far as possible for the user. The user, ideally the domain expert, using the machine learning models for holistic asset monitoring, should be enabled to carry out all sensor and model related tasks using the web dashboard (Fig 1). The installation of the sensor should also be as simple as possible. An exact sensor position can be specified on an asset, while an exact sensor rotation is challenging. Consequently, a virtual sensor alignment is required.

\section{Proposed Machine Learning Pipeline}
The proposed machine learning pipeline consists of 1) virtual sensor alignment 2) employment of a general classification model learning from all classes 3) post-processing of the classification output 4) specific model learning from healthy class and 5) the model voting (Fig 2). Virtual sensor alignment ensures the rotational invariance of the sensor by remapping the sensor from the sensor coordinate system which can differ between sensors into a unified virtual one. The generic classification model is trained with all classes and is assumed to be trained long-term on an increased number of data from various different assets. An autoencoder can be employed as specific detection model and only trained on healthy data. As this data is relatively inexpensive to collect, the autoencoder is assumed to be trained individually for each asset. Two different approaches were explored, a) a feature-based ANN and b) an end-to-end approach ROCKET. From the resulting Logits the probabilities of both approaches were derived and smoothed with a moving average filter. Finally, voting between generic classification and specific detection model was applied.

\begin{figure}[ht]
	\centerline{\includegraphics[width=0.45\textwidth]{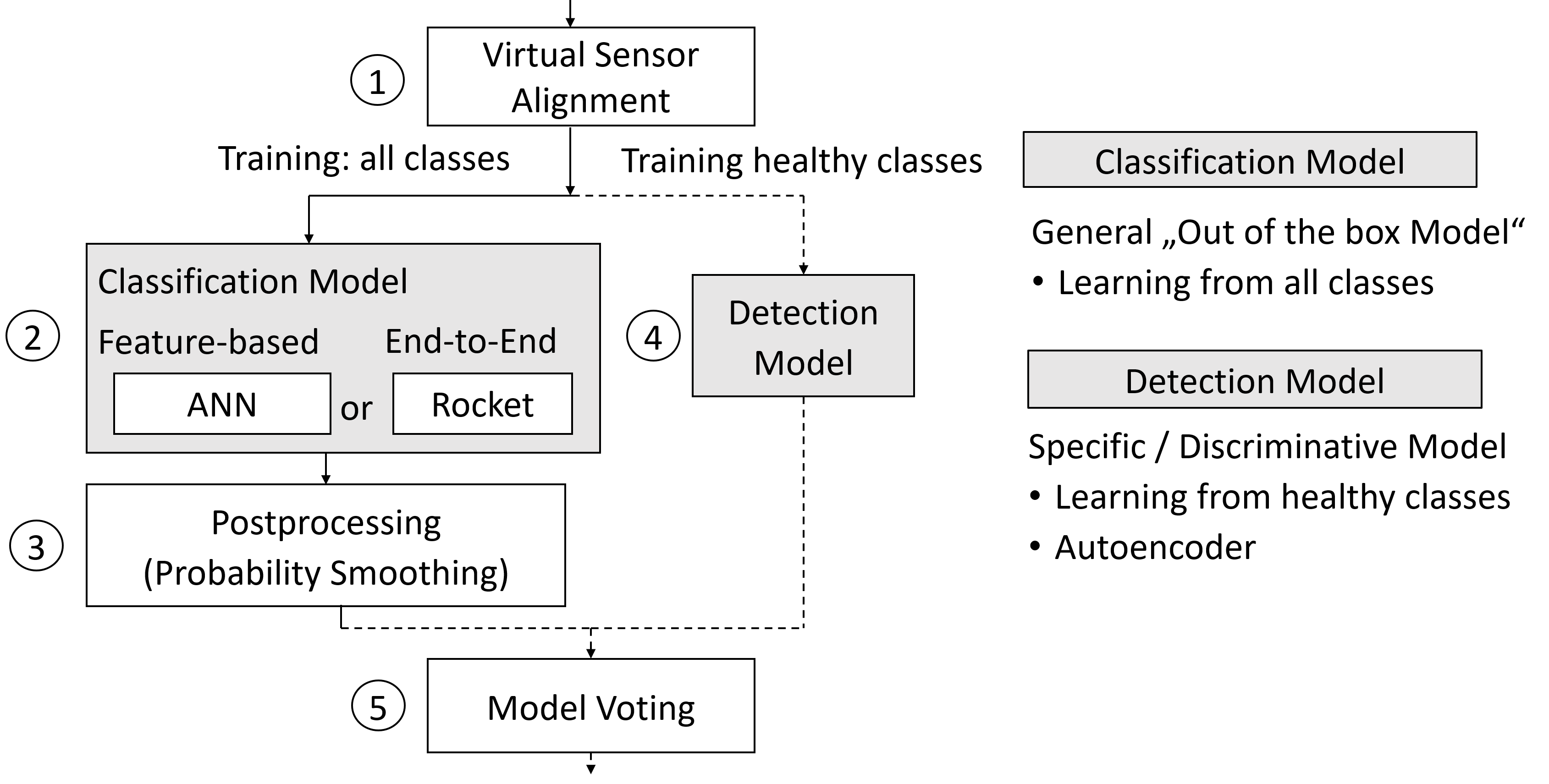}}
	\caption{Proposed machine learning pipeline}
	\label{fig1}
\end{figure}

\textit{1) Virtual Sensor Alignment.}
 To increased robustness and transferability, rotational invariance of the machine learning model input data is suggested. We propose remapping the data from sensor coordinate system $s_i$ into the world coordinate system $w_i$ using the Kabsch algorithm \cite{Markley1993} minimizing the following loss function L which is solved for the rotation matrix C (to align $s_i$ to $w_i$):

\begin{equation}
 L(C) = \sum_{i=1}^n { || S_{ i }-C_{w_{i}  } ||}^{ 2 }
 \end{equation}

\textit{2) Classification Model.}

\textit{a) Feature-based Approach: ANN.} The ANN approach employed a sliding window on the time series data of the virtual axes, followed by the extraction of the Mel-frequency cepstral coefficients (mfccs), minority oversampling and normalization (Fig 3). The resulting 20 coefficients were feed into the ANN.
 
\textit{Data Augmentation and Feature Selection.} Besides more widely used as features for audio classification, we suggest mfccs for vibration data as both having strong relation due to air-borne and structure-borne sound transmission \cite{Cremer2005}. The mfccs represent the power spectrum of the short-term Fourier transform on a nonlinear frequency scale inspired by human biology, uniformly spaced below 1 kHz and logarithmic scale above 1 kHz. Two different data augmentation techniques are suggested, a) sliding window on time series data and b) minority oversampling on the extracted features. Sliding window: The $1$x$512$ input values are artificially increased towards $16$x$256$ with window size of $256$ and on offset of 16. Minority oversampling: Collecting data from anomaly classes is challenging and often has a detrimental effect on an industrial asset. Therefore, synthetic minority oversampling technique (SMOTE) \cite{Chawla2002} of the mfccs features is proposed in the pipeline.

\begin{figure}[ht]
	\centerline{\includegraphics[width=0.45\textwidth]{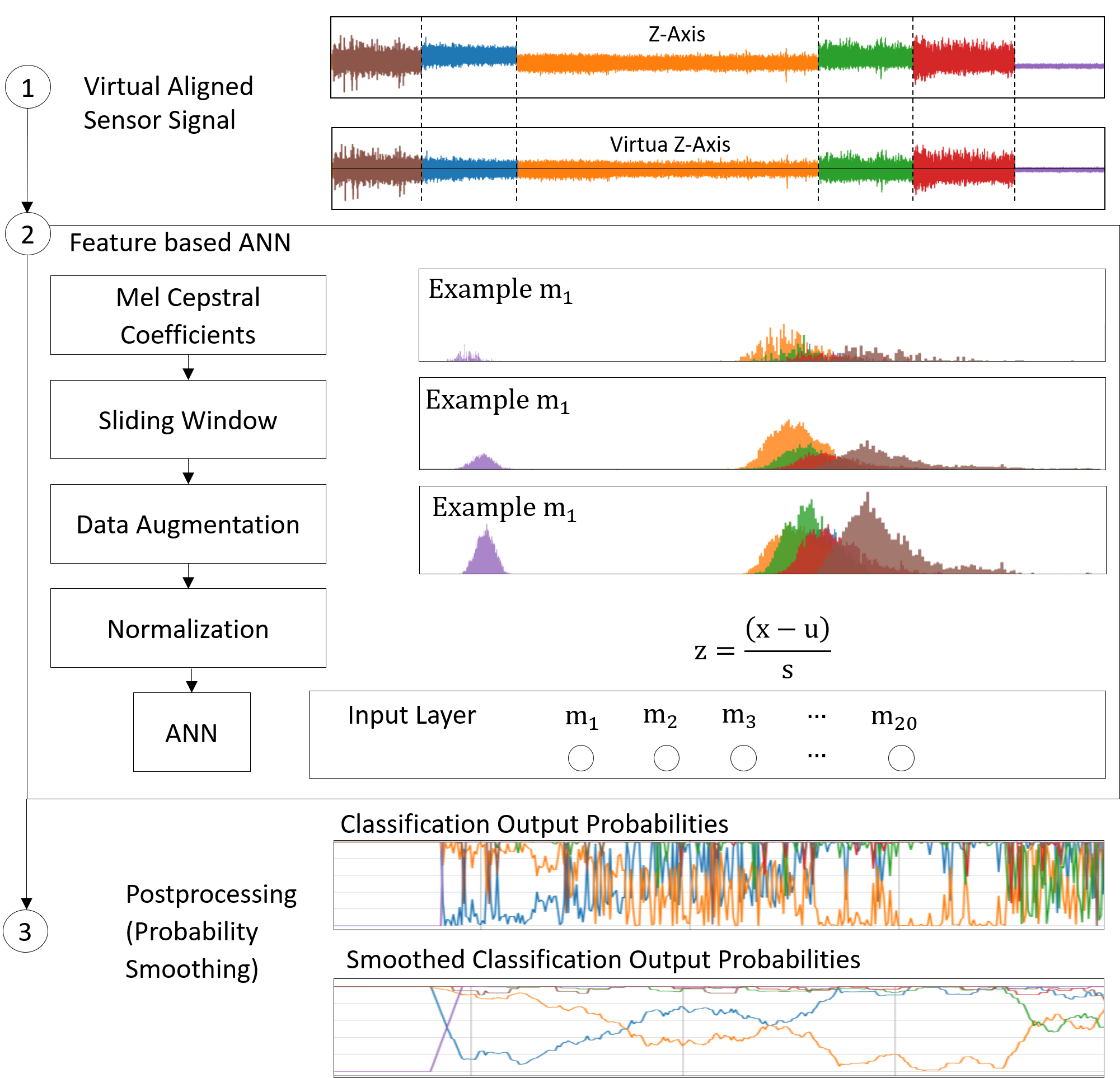}}
	\caption{Data visualization for 1) virtual Sensor alignment, 2) feature-based ANN and 3) probability smoothing of the proposed machine learning pipeline}
	\label{fig3}
\end{figure}
 
\textit{Hyperparameter Consideration.} When designing the architecture for an ANN, a variety of parameters can be tuned. The art is to find the right combination for these parameters to achieve the highest accuracy and lowest loss. Therefore, TALOS \cite{TALOS} was employed for the hyperparameter search. The framework allows to randomly sample from a given grid of hyperparameters and train the respective networks to support the user finding the best combination. The following parameters resulted from the framework:  The respective $\alpha$: 0.001, $\beta$: 0.9 $\beta_1$: 0.9, $\beta_2$:0.999, $\epsilon$: \num{e-8} for Adam optimization, $\#$ layers: 2, $\#$ hidden units: 64 (each hidden layer), relu activation function, mini batch size: 16 and a SoftMax output layer. The TALOS framework, however, does not allow to design best network parameters for training and test data from different distribution. Consequently, this resulted in a very small bias, but high variance as the model was overfitting to the training data. Therefore, we employed dropout of 40$\%$ and early stopping after 30 epochs. The resulting mccfs were normalized between [-1,1] based on a transformation resulting in a Gaussian distribution before fed into the Neural Network.

\subsection{End-to-End-Approach: ROCKET.}
The end-to-end machine learning model ROCKET \cite{Dempster2020} transforms the virtual aligned sensor signal using a large number of convolutional kernels. The convolutional kernels are randomly created varying length, weights, bias, dilation, and padding. The transformed features were used to train a linear ridge classifier. The classifier relies on cross-validation and L2 regularization to avoid overfitting to the training data, accepting a defined bias to reduce the variance. The combination of ROCKET and regression forms, in effect, a single-layer convolutional neural network with random kernel weights, where the transformed features form the input for a trained SoftMax layer \cite{Dempster2020}. 

\textit{3) Output Probability Considerations.}
As the network prediction only takes the current point in time into account, which have been observed to fluctuate quite strongly (Fig 3, Classification Output Probabilities, Fig 4, (2)), we employed a filter to smooth the probabilities. In this case a moving average filter [size: 15] was employed (Fig 3, Smoothed Classification Output Probabilities, Fig 4, (3)). Since it is a SoftMax layer, raw probabilities can be acquired similar to the ANN and ROCKET approach.

\textit{4) Detection Model.}
It can be assumed that an unlimited amount of healthy training data is available for an industrial asset. This data can be used for asset specific anomaly detection to increase the overall robustness of our ML pipeline. Therefore, we trained an unsupervised machine learning model to detect the anomalies in our dataset by using an autoencoder architecture based fully connected deep neural network (DNN). The autoencoder learns to reconstruct the input for healthy sensor data, as it was trained to do so, but will fail to reconstruct anomaly data. The reconstruction error, the error between input and output signal, was used as an anomaly score. The threshold was calculated as mean + standard deviation of the reconstruction error of healthy data. The DNN architecture consists of five fully connected layers with the respective number of neurons per layers $\#$512 input $\#$256 encoding 126 bottleneck $\#$256 decoding $\#$512 output; $tanh$ was employed as activation function.

\textit{5) Model Voting.}
The model voting was done in a way, that the autoencoder always overruled the classification results and a non-healthy class had to change to the anomaly class with the highest probability and vice versa, if the autoencoder predicted the other class (Fig 4, (5)).

\begin{figure}[ht]
	\centerline{\includegraphics[width=0.45\textwidth]{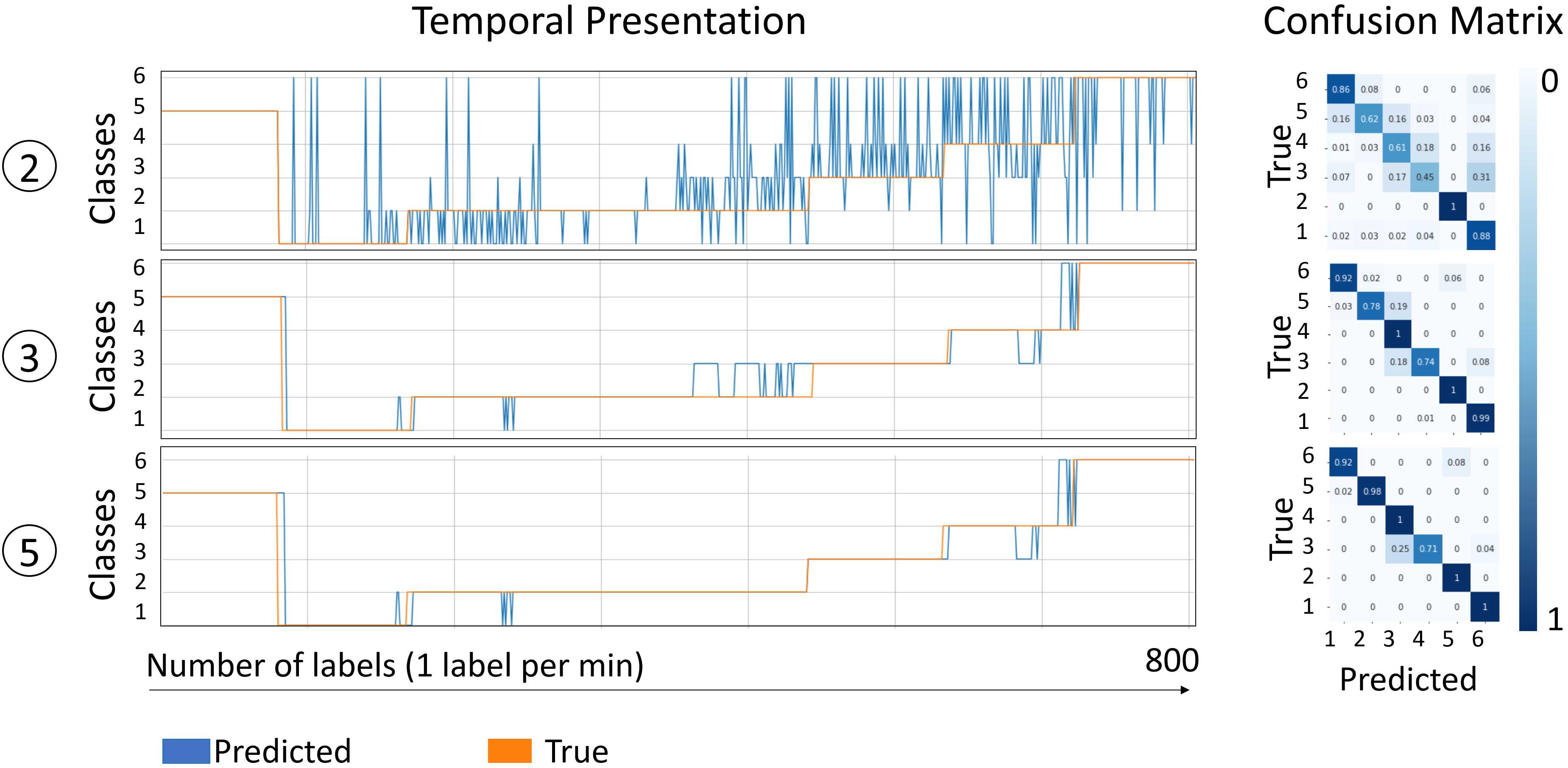}}
	\caption{Prediction results at different steps within the proposed machine learning pipeline. Training data 1; example On/Off, feature-based ANN approach}
	\label{fig3}
\end{figure}

\section{Measurement Setup and Data Set Description}
All measurements were done in a laboratory test bench. As an industrial asset, a centrifugal pump motor combination was selected. Two series of measurements with each one training data set and four test data sets were created. The first series of measurements was created to examine the robustness of a model during maintenance work or sensor battery change within the same pump, and the second to examine how well the models can be transferred to other pumps of the same construction.

Measurement Series I (within pump):
\begin{itemize}
\item Training Set I: Reference
\item Test Set 1 Asset was turned off and cooled down 
\item Test Set 2: Sensor was removed and reattached
\item Test Set 3: Screws were removed and reattached
\item Test Set 4: Asset was completely dismantled and rebuilt
\end{itemize}

Measurement Series II (between pumps): 
\begin{itemize}
\item Training Set II: Pump X
\item Test Set 5: Pump I
\item Test Set 6: Pump II
\item Test Set 7: Pump III
\item Test Set 8: Pump IV
\end{itemize}

The data sets consisted of six classes, i.e., three healthy operational conditions and three non-healthy operational conditions. 

Healthy (Normal) Operational Conditions: 
\begin{itemize}
\item Class 1, normal load (flow rate of \SI[per-mode=reciprocal]{50}{\cubic\metre\per\hour})
\item Class 2, partial load (flow rate of 12.5$-$ \SI[per-mode=reciprocal]{37.5}{\cubic\metre\per\hour})
\item Class 6, idle state (flow rate of \SI[per-mode=reciprocal]{0}{\cubic\metre\per\hour})
\end{itemize}

Non-Healthy (Anomaly) Conditions:
\begin{itemize}
\item Class 3, dry running pump (flow rate of \SI[per-mode=reciprocal]{0}{\cubic\metre\per\hour})
\item Class 4, hydraulic blockage (flow rate of \SI[per-mode=reciprocal]{0}{\cubic\metre\per\hour})
\item Class 5, cavitation (nominal point 50 \SI[per-mode=reciprocal]{60}{\cubic\metre\per\hour}, net positive suction head (NPSH) of -0.8)
\end{itemize}
Each class was measured at least 30 min per pump, corresponding to 30 labels. The measurements were acquired on different days and even months, but according to the same measurement protocol. As anomalies were introduced manually, there are deviations within the classes. This, however, can be considered as a real-world problem, as those data are also expected to be broader distributed. 

\section{Machine Learning Experiments }
We aimed to evaluate the impact and challenges of production divergence and maintenance in the context of robustness and transferability systematically.
Therefore, we performed an initial baseline experiment where training and test set were derived from the same data sets for Training Data I and II.
Subsequently, we performed systematic experiments: first, we trained one Model I on Training Set I and Model II on Training Set I and II. The trained models were tested on all eight test sets. To compare different pre-processing algorithms and machine learning models, we evaluated all experiments with an ANN and ROCKET. Both methods were evaluated without virtual sensor alignment (M), with virtual sensor alignment (V+M), with subsequent probability smoothing (V+M+S) and subsequent employing of the autoencoder (V+M+S+A). The ideal criteria for using an autoencoder, that training and test set come from the same pump where no maintenance work has been performed on either the pump or the sensor, was met only by test set 1.

Hypothesis were tested with paired or unpaired t-test (depending on the comparison). 
 
\section{Experimental Results and Conclusion}
The baseline classification accuracy was quite high ($>$0.972) for both data sets (table 1). ROCKET showed slightly higher performance and was essential faster during the training (but not prediction). 

\begin{table}[!htbp]
\centering
\scriptsize
\caption{Accuracy, Baseline Experiments}
\begin{tabular}{lllll} 
\toprule
                & \multicolumn{2}{l}{ANN} & \multicolumn{2}{l}{ROCKET}  \\ 
\cline{2-5}
                & M     & V+M             & M     & V+M                 \\ 
\hline
Training Data I & 0.976 & 0.978           & 0.983 & 0.973               \\
Training Data II & 0.972 & 0.973           & 0.986 & 0.986               \\ 
\hline
\multicolumn{5}{l}{M-model; V-virtual sensor alignment}                 \\
\bottomrule
\end{tabular}
\end{table}

Results for the robustness and transferability experiments are shown in table 2. 

\subsubsection{Feature-based ANN vs ROCKET approach}
The average accuracy over all robustness and transferability experiments was higher in the ROCKET approach (0.862$\pm$0.116) compared to the feature-based ANN (0.857$\pm$0.078), the difference was statistically significant (p$=$0.002, paired t-test) (Model Voting was excluded as available for one experiment only). Including only results with respective pre- and post-processing steps (virtual sensor alignment and probability smoothing) the differences vanished: ROCKET approach (0.909$\pm$0.104) compared to ANN (0.908$\pm$0.063) (p$=$0.965, paired t-test).

\subsubsection{Same vs Different Measurement Series}
The impact of using the same measurement series vs a different one is clearly high, with an average accuracy of both models 0.978$\pm$0.005 on the baseline experiments vs 0.829$\pm$0.067 on the transferability and robustness experiments without pre- and post-processing (unpaired t-test: $<$0.001). Pre- and post-processing, however, decreased the difference for average results of both models (0.908$\pm$0.068), but remained significant (unpaired t-test: $<$0.015). 

\subsubsection{Same vs Different Pump Series}
Average accuracy classification accuracy was notably higher in same pumps (0.848$\pm$0.062) vs different pumps (0.809$\pm$0.069) but did not reach statistically significance (p$=$0.055, unpaired t-test). The difference was somewhat lower (but still notable even though significance vanished even more), employing pre- and post-processing methods: same pumps (0.924$\pm$0.058) vs different pumps (0.892$\pm$0.105), (p$=$0.154, unpaired t-test).  

\subsubsection{Sensor Alignment} 
Without post-processing, the sensor alignment had a
slightly negative impact with averaged model results
without (0.829$\pm$0.069) vs with (0.811$\pm$0.096)
virtual sensor alignment (p$=$0.05); with post-processing
the relationship turned around model results without (0.889$\pm$0.106) vs with (0.908$\pm$0.086) virtual
sensor alignment (p$=$0.06). Note: Sensor position was
attached according to the same protocol and sensor was not rotated intentionally.

\begin{table}[!htbp]
\centering
\scriptsize
\caption{Accuracy, Robustness and Transferability Experiments}
\begin{tabular}{lllllllll} 
\toprule
\multicolumn{9}{l}{\textbf{Feature-based Approach: ANN}}                                                      
\\ 
\hline
& \multicolumn{4}{c}{Within Pumps} &\multicolumn{4}{c}{Between Pumps}
                                                                                                                                                                                                                                                                    \\

                 & \begin{tabular}[c]{@{}l@{}}T1\end{tabular} & \begin{tabular}[c]{@{}l@{}}T2\end{tabular} & \begin{tabular}[c]{@{}l@{}}T3\end{tabular} & \begin{tabular}[c]{@{}l@{}}T4\end{tabular} &T5& T6&T7&T8                                                   \\
\hline
\multicolumn{9}{l}{Model I (Training Data I)}                                                                                                                                                                                                                                                                                                                      \\
\hline
\textbf{\hspace{0.36em} M}      & 0.842                                            & 0.804                                             & 0.884                                             & 0.788                                    & 0.809          & 0.755          & 0.816          & 0.825                                                               \\
\textbf{\hspace{0.36em} M}S   & 0.917                                            & 0.929                                             & 0.951                                             & \textbf{0.800}                                       & \textbf{0.819} & 0.822          & \textbf{0.992} & 0.905                                                               \\
V\textbf{M}     & 0.852                                            & 0.822                                             & 0.888                                             & 0.616                                             & 0.822          & 0.783          & 0.843          & 0.807                                                               \\
V\textbf{M}S   & 0.928                                            & \textbf{0.931}                                   & \textbf{0.974}                                & 0.788                                    & \textbf{0.819} & \textbf{0.953} & \textbf{0.992} & \textbf{0.921}                                                      \\
V\textbf{M}SA & \textbf{0.985}                                   &                                                 &                                                    &                                                    &                &                &                &                                                                     \\
\hline
\multicolumn{9}{l}{Model II (Training Data I + Training Data II)}                                                                                                                                                                                                                                                                                                                      \\
\hline
\textbf{\hspace{0.36em} M}      & 0.861                                            & 0.801                                             & 0.876                                             & 0.721                                             & 0.811          & 0.772          & 0.825          & 0.922                                                               \\
\textbf{\hspace{0.36em} M}S   & 0.928                                            & \textbf{0.904}                                 & \textbf{0.938}                                    & 0.804                                             & 0.819 & 0.819          & \textbf{0.990} & \textbf{0.975}                                                      \\
V\textbf{M}     & 0.849                                            & 0.787                                             & 0.849                                             & 0.637                                             & \textbf{0.821}         & 0.818          & 0.796          & 0.884                                                               \\
V\textbf{M}S   & 0.913                                            & 0.891                                             & 0.912                                             & \textbf{0.813}                                 & 0.819 & \textbf{0.982} & 0.912          & 0.972                                                               \\
V\textbf{M}SA & \textbf{0.980}                                   &                                                  &     
                                             &                                              &                &                &                &                                                                     \\
                 &                                                  &                                                   &                                                   &                                                   &                &                &                &                                                                     \\
\multicolumn{9}{l}{\textbf{End-End-based Approach: ROCKET}}                                                                                                                                                                                                                                                                                                                      \\ 
\hline
\multicolumn{9}{l}{Model I (Training Data I)}                                                                                                                                                                                                                                                                                                                      \\ 
\hline
\textbf{\hspace{0.36em} M}                & 0.893                                            & 0.850                                             & 0.934                                             & 0.760                                             & 0.779          & 0.778          & 0.661          & 0.725                                                               \\
\textbf{\hspace{0.36em} M}S             & 0.920                                            & \textbf{0.936}                                    & 0.994                                             & 0.646                                             & 0.795          & 0.923          & 0.523          & \textbf{0.806}                                                               \\
V\textbf{M}              & 0.882                                            & 0.846                                             & 0.944                                             & 0.835                                             & \textbf{0.822} & 0.779          & 0.518          & 0.658                                                               \\
V\textbf{M}S            & 0.917                                            & 0.934                                             & \textbf{0.995}                                    & \textbf{0.939}                                   & 0.819          & \textbf{0.951} & \textbf{0.622} & 0.742                                                      \\
V\textbf{M}SA          & \textbf{0.994}                                   &       
                                            &     
                                            &                                                &                &                &                &                                                                     \\ 
\hline
\multicolumn{9}{l}{Model II (Training Data I + Training Data II)}                                                                                                                                                                                                                                                                                                                      \\ 
\hline
\textbf{\hspace{0.36em} M}                & 0.910                                            & 0.895                                             & 0.952                                             & 0.801                                             & 0.782          & 0.909          & 0.840          & 0.933                                                               \\
\textbf{\hspace{0.36em} M}S             & 0.991                                            & 0.975                                             & 0.993                                    & \textbf{0.904}                                  & 0.803          & 0.984          & \textbf{0.987}         & 0.966                                                               \\
V\textbf{M}              & 0.909                                            & 0.887                                             & \textbf{0.944}                                             & 0.637                                             & \textbf{0.821}          & 0.887          & 0.848          & 0.876                                                               \\
V\textbf{M}S            & 0.990                                            & \textbf{0.986}                                    & 0.992                                             & 0.884                                             & 0.819 & \textbf{0.998} & \textbf{0.987} & \textbf{0.967}                                                      \\
V\textbf{M}SA          & \textbf{0.998}                                   &        
                                           &     
                                              &     
                                              &                &                &                &                                                                     \\ 
\hline
\multicolumn{9}{l}{\begin{tabular}[c]{@{}l@{}}Rem-Removal and Installation; T-Test; M-Model; V-Virtual Sensor Alignment\\
S-Smoothing of Probabilities; A-Autoencoder; Training data 1: smaller model;\\
Training data 2: larger, more diverse model \end{tabular}}    \\
\bottomrule
\end{tabular}
\end{table}

\subsubsection{Probability Smoothing} 
The impact of probability smoothing is clearly high on all averaged model accuracy before smoothing (0.812$\pm$0.063) and after smoothing (0.901$\pm$0.066) and highly significant (p$<$0.001, paired-t-test). 

\subsubsection{Variation in Training Data} There was no worsening nor any improvement when the training data was either only acquired from the same pump (training data 1: 0.873$\pm$0.087) or from the same and a different pump (training data 1: 0.879$\pm$0.091) (p$<$0.308, paired-t-test). However, when training and test measurements were acquired from different pumps, the more complex training data had significantly positive impact on the accuracy: average accuracy training data 1 (0.800$\pm$0.087) vs training data 2 (0.879$\pm$0.091) (p$<$0.001, paired-t-test).

\subsubsection{Impact of Model Voting} The model voting employing the anomaly detection results of an autoencoder, trained on normal data only, increased the classification accuracy in all cases explored (test set 1 only): without (0.937$\pm$0.036) with (0.989$\pm$0.008) (p$<$0.042, paired-t-test).

\section{Conclusion and Future Work}
This work is limited to a single asset type and a single environment. The influence of these factors may be investigated in future studies. A limitation of the study is the small numbers of models compared. However, the aim was to provide some initial guideline for the overall machine learning pipeline. 

First, we endorse a fixed sensor position. The virtual sensor alignment showed a small impact, however, in all measurements, the sensor position was attached according to the same protocol and not rotated intentionally. Such an exact placement of the sensor may not always be the case in a real-world scenario. Therefore, the virtual alignment is recommended. Additionally, we suggest the employment of probability smoothing as post-processing. The approach can be used independent of the model, as the raw probabilities can be acquired from the Logits in each model using a SoftMax layer for classification. Additionally, we recommend increasing the complexity of the training data. Further, a specific autoencoder, trained on easy-to-collect healthy data combined with the classification results is proposed.

\nocite{*}
\bibliography{robust_transfer}{}
\bibliographystyle{IEEEtran}

\end{document}